\documentclass[twoside,twocolumn,9pt]{article}
\usepackage{extsizes}
\usepackage[super,sort&compress,comma]{natbib} 
\usepackage[version=3]{mhchem}
\usepackage[left=1.5cm, right=1.5cm, top=1.785cm, bottom=2.0cm]{geometry}
\usepackage{balance}
\usepackage{mathptmx}
\usepackage{sectsty}
\usepackage{graphicx}
\usepackage{lastpage}
\usepackage[format=plain,justification=justified,singlelinecheck=false,font={stretch=1.125,small,sf},labelfont=bf,labelsep=space]{caption}
\usepackage{float}
\usepackage{fancyhdr}
\usepackage{fnpos}
\usepackage[english]{babel}
\addto{\captionsenglish}{%
  
}
\usepackage{array}
\usepackage{droidsans}
\usepackage{charter}
\usepackage[T1]{fontenc}
\usepackage[usenames,dvipsnames]{xcolor}
\usepackage{setspace}
\usepackage[compact]{titlesec}
\usepackage{hyperref}
\usepackage{epstopdf}%This line makes .eps figures into .pdf - please comment out if not required.
\definecolor{cream}{RGB}{222,217,201}

\usepackage[framemethod=TikZ]{mdframed} %背景框
\usepackage{xcolor}
\definecolor{hlcolor}{RGB}{255,255,0} % 黄色高亮

\usepackage{hyperref} % 加载 hyperref 宏包

\usepackage{tikz} %图片背景高亮

\newmdenv[
	backgroundcolor=yellow!30,
	hidealllines=true,
	innerleftmargin=0pt,
	innerrightmargin=0pt,
	innertopmargin=0pt,
	innerbottommargin=0pt,
]{highlight}

\begin{document}

\pagestyle{fancy}
\thispagestyle{plain}
\fancypagestyle{plain}{
%%%HEADER%%%
\renewcommand{\headrulewidth}{0pt}
}
%%%END OF HEADER%%%

%%%PAGE SETUP - Please do not change any commands within this section%%%
\makeFNbottom
\makeatletter
\renewcommand\LARGE{\@setfontsize\LARGE{15pt}{17}}
\renewcommand\Large{\@setfontsize\Large{12pt}{14}}
\renewcommand\large{\@setfontsize\large{10pt}{12}}
\renewcommand\footnotesize{\@setfontsize\footnotesize{7pt}{10}}
\makeatother

\renewcommand{\thefootnote}{\fnsymbol{footnote}}
\renewcommand\footnoterule{\vspace*{1pt}% 
\color{cream}\hrule width 3.5in height 0.4pt \color{black}\vspace*{5pt}} 
\setcounter{secnumdepth}{5}

\makeatletter 
\renewcommand\@biblabel[1]{#1}            
\renewcommand\@makefntext[1]% 
{\noindent\makebox[0pt][r]{\@thefnmark\,}#1}
\makeatother 
\renewcommand{\figurename}{\small{Fig.}~}
\sectionfont{\sffamily\Large}
\subsectionfont{\normalsize}
\subsubsectionfont{\bf}
\setstretch{1.125} %In particular, please do not alter this line.
\setlength{\skip\footins}{0.8cm}
\setlength{\footnotesep}{0.25cm}
\setlength{\jot}{10pt}
\titlespacing*{\section}{0pt}{4pt}{4pt}
\titlespacing*{\subsection}{0pt}{15pt}{1pt}
%%%END OF PAGE SETUP%%%

%%%FOOTER%%%
\fancyfoot{}
%\fancyfoot[LO,RE]{\vspace{-7.1pt}\includegraphics[height=9pt]{head_foot/LF}}
%\fancyfoot[CO]{\vspace{-7.1pt}\hspace{13.2cm}\includegraphics{head_foot/RF}}
%\fancyfoot[CE]{\vspace{-7.2pt}\hspace{-14.2cm}\includegraphics{head_foot/RF}}
\fancyfoot[RO]{\footnotesize{\sffamily{1--\pageref{LastPage} ~\textbar  \hspace{2pt}\thepage}}}
\fancyfoot[LE]{\footnotesize{\sffamily{\thepage~\textbar\hspace{2pt} 1--\pageref{LastPage}}}}
%3.45cm
\fancyhead{}
\renewcommand{\headrulewidth}{0pt} 
\renewcommand{\footrulewidth}{0pt}
\setlength{\arrayrulewidth}{1pt}
\setlength{\columnsep}{6.5mm}
\setlength\bibsep{1pt}
%%%END OF FOOTER%%%

%%%FIGURE SETUP - please do not change any commands within this section%%%
\makeatletter 
\newlength{\figrulesep} 
\setlength{\figrulesep}{0.5\textfloatsep} 

\newcommand{\topfigrule}{\vspace*{-1pt}% 
\noindent{\color{cream}\rule[-\figrulesep]{\columnwidth}{1.5pt}} }

\newcommand{\botfigrule}{\vspace*{-2pt}% 
\noindent{\color{cream}\rule[\figrulesep]{\columnwidth}{1.5pt}} }

\newcommand{\dblfigrule}{\vspace*{-1pt}% 
\noindent{\color{cream}\rule[-\figrulesep]{\textwidth}{1.5pt}} }

\makeatother
%%%END OF FIGURE SETUP%%%

%%%TITLE, AUTHORS AND ABSTRACT%%%

\twocolumn[
  \begin{@twocolumnfalse}

\begin{tabular}{m{1cm} p{16cm} m{1cm}}

& \noindent\LARGE{\textbf{Toward Automated Simulation Research Workflow through LLM Prompt Engineering Design}} & \\%Article title goes here instead of the text "This is the title"
\vspace{0.3cm} & \vspace{0.3cm}  & \vspace{0.3cm}\\

& \noindent\large{Zhihan Liu,\textit{$^{a}$} Yubo Chai,\textit{$^{a}$} Jianfeng Li,$^{\ast}$\textit{$^{a}$}} & \\%Author names go here instead of "Full name", etc.$^{\ddag}$
\vspace{0.3cm} & \vspace{0.3cm} & \vspace{0.3cm}\\

& \noindent\normalsize{
   ABSTRACT: The advent of Large Language Models (LLMs) has created new opportunities for the automation of scientific research spanning both experimental processes and computational simulations. This study explores the feasibility of constructing an autonomous simulation agent (ASA) powered by LLMs through prompt engineering and automated program design to automate the entire simulation research process according to a human-provided research plan. This process includes experimental design, remote upload and simulation execution, data analysis, and report compilation. Using a     well-studied simulation problem of polymer chain conformations as a test case, we assessed the long-task completion and reliability of ASAs powered by different LLMs, including GPT-4o, Claude-3.5, etc. Our findings revealed that ASA-GPT-4o achieved near-flawless execution on designated research missions, underscoring the potential of methods like ASA to achieve automation in simulation research processes to enhance research efficiency. The outlined automation can be iteratively performed for up to 20 cycles without human intervention, illustrating the potential of ASA for long-task workflow automation. Additionally, we discussed the intrinsic traits of ASA in managing extensive tasks, focusing on self-validation mechanisms, and the balance between local attention and global oversight.
   
} & \\%The abstrast goes here instead of the text "The abstract should be..."

\end{tabular}

 \end{@twocolumnfalse} \vspace{0.6cm}
]
%%%END OF TITLE, AUTHORS AND ABSTRACT%%%

%%%FONT SETUP - please do not change any commands within this section
\renewcommand*\rmdefault{bch}\normalfont\upshape
\rmfamily
\section*{}
\vspace{-1cm}

%%%FOOTNOTES%%%

\footnotetext{\textit{$^{a}$ The State Key Laboratory of Molecular Engineering of Polymers, The Research Center of AI for Polymer Science Department of Macromolecular Science, Fudan University, Shanghai 200433, China; E-mail: $\ast$ lijf@fudan.edu.cn}}

%$\ddag$ lyx@fudan.edu.cn
%Please use \dag to cite the ESI in the main text of the article.
%If you article does not have ESI please remove the the \dag symbol from the title and the footnotetext below.
%\footnotetext{\dag~Electronic Supplementary Information (ESI) available: [details of any supplementary information available should be included here]. See DOI: 00.0000/00000000.}
%additional addresses can be cited as above using the lower-case letters, c, d, e... If all authors are from the same address, no letter is required

%\footnotetext{\ddag~Additional footnotes to the title and authors can be included \textit{e.g.}\ `Present address:' or `These authors contributed equally to this work' as above using the symbols: \ddag, \textsection, and \P. Please place the appropriate symbol next to the author's name and include a \texttt{\textbackslash footnotetext} entry in the the correct place in the list.}

%%%END OF FOOTNOTES%%%
\section*{INTRODUCTION}
%%%MAIN TEXT%%%%

In recent years, the application of artificial intelligence (AI) in scientific research has gained significant traction. AI technologies, through data analysis, pattern recognition, and predictive capabilities, offer researchers unprecedented support, enhancing both efficiency and accuracy.

AI is utilized in chemical synthesis, drug design, and materials science, accelerating the achievement of valuable research outcomes. Meta's Galactica supports researchers in organizing information, inferring knowledge, and writing, highlighting the potential of AIs in scientific contexts.\cite{Taylor2022} DeepMind's AlphaFold algorithms enable the prediction of structures for biological molecules, from proteins and nucleic acids to larger and more complex complexes.\cite{Evans2021,Abramson2024} Machine Learning methods not only enhance the understanding and predictive capabilities of complex systems but also offer new approaches to tackling problems that are difficult for traditional computational methods to handle.\cite{JiangJ2024a,JiangJ2024b,LiJF2019,LiJF2023,SunZY2024a,SunZY2024b} Integrating machine learning with automated experimental techniques enables faster experimental design and data analysis, leading to quicker discoveries of new materials and improvements in existing material properties.\cite{Himanen2019,Wang2023}

Among various AI technologies, tools based on large language models (LLMs) such as GPT and Claude have demonstrated superior capabilities in generating and understanding complex texts across a range of natural language processing tasks.\cite{OpenAI2023,Anthropic2023} In addition, LLMs can also generate corresponding code based on text descriptions, significantly simplifying programming tasks.\cite{Xu2022, Hou2024} Research shows that LLMs excel in handling complex logic, code completion, and debugging, making them valuable assistants in software development and data analysis.\cite{Burstein2019,JMLR2020,Floridi2020,Chowdhery2022PaLMSL,Thoppilan2022LaMDALM} In the field of scientific research, the application prospects of LLMs are increasingly evident, gradually transforming traditional research methods and providing scientists with unprecedented tools and methodologies.\cite{ramesh2021zero,bommasani2022opportunitiesrisksfoundationmodels,Chang2024,KASNECI2023} For example, integrating LLMs with specialized chemical tools and databases can extend their capabilities beyond general language tasks.\cite{Brown2019,Stokes2020,Boiko2023,Bran2024,Kang2024} Researchers have also utilized LLMs to automate literature reviews\cite{Guo2024, Bolanos2024, Lozano2024, Dagdelen2024} and data analysis\cite{Fink2023, Ma2023, Peng2024} and assist in experimental design and species screening \cite{Li2024, Subba2024, Liu2023, Kang2024, Madani2023} to improve research efficiency. Additionally, some studies highlighted the crucial role of prompt engineering in guiding models to produce accurate and useful outputs. Techniques such as Chain of Thought, ReAct, and Tree of Thoughts help structure the model's reasoning process, reduce errors, and improve the quality of generated results.\cite{Bran2024,Jablonka2024,Schulhoff2024,Zhou2024,Yao2023,Wei2022,Kojima2022} These advancements allow researchers to dedicate more time to innovative thinking and theoretical exploration rather than repetitive and inefficient experimental operations.

The aforementioned studies show that continuous human-AI interaction can accelerate various stages of scientific research. However, the current model of these studies mainly involves interaction between humans and AI. By utilizing the robust capabilities of large language models (LLMs), we aspire to gradually reduce human involvement in the research process, aiming to enable AI to independently complete end-to-end workflows, thus further enhancing research efficiency.

Taking simulation research as an example, in traditional simulation research processes, researchers develop a research plan (RP) that specifies the simulation methods to be used, how to write programs or use simulation software, how to build models, and which parameters to test. Following the establishment of the RP, a significant amount of experimentation is carried out to gather data, which are then analyzed to draw scientific conclusions. After the RP is determined, the subsequent steps are largely standardized and present opportunities for automation. Therefore, we intend to test whether, by providing a detailed textual RP, the advanced capabilities of LLMs can allow AI to successfully and automatically complete an entire simulation research process without human intervention. This approach could streamline the research workflow and lead to more efficient outcomes.

With their outstanding language understanding and code generation capabilities, LLMs can produce rich and varied content based on prompts. However, they cannot guarantee the accuracy of the generated code in a single attempt, and due to the length and complexity of tasks described in the RP, LLMs may overlook subtasks. Additionally, as language models, LLMs do not have the ability to control local computers to execute programs. To address these challenges and achieve an automated workflow, we developed an automated system that can identify and run the code generated by the LLM on a local machine. We also used prompt engineering to guide the LLM in producing structured and accurate responses. Through multiple rounds of interaction, we were able to instruct the LLM to debug program errors, check the completeness of task execution, and make the necessary adjustments. This approach resulted in the creation of an end-to-end Autonomous Simulation Agent (ASA). As demonstrated in the demo video, simply submitting an RP txt file via the command ``do AI4PS\_PRJ1.txt'' to ASA allows it to execute all subsequent tasks without any human intervention; ASA can even automatically upload programs to a remote server and perform a series of simulations---something that cannot be achieved solely through interactions with a web-based LLM.

We chose a problem in polymer physics with known outcomes to evaluate the feasibility and reliability of ASA in automating research workflows. Detailed RPs were provided to the ASA, which involved the AI fully autonomously coding simulation programs, remotely uploading and executing simulations, organizing data, plotting results, and writing scientific reports. We also designed several evaluation criteria to determine if the ASA could correctly and completely perform each step according to RP without human intervention.

\begin{figure}[h!]
	\centering
	\includegraphics[width=0.5\textwidth]{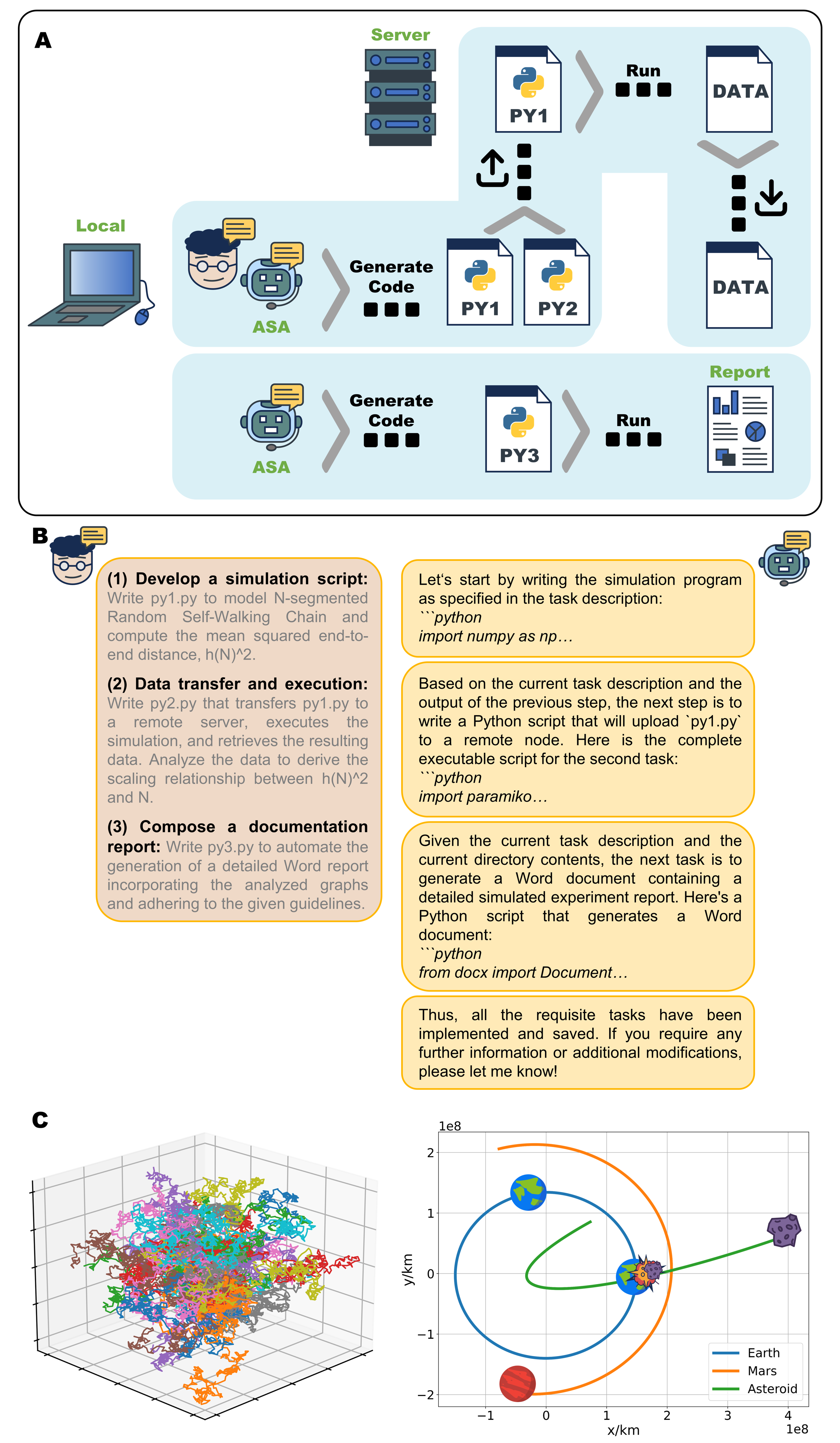}
	\captionsetup{justification=justified} % 明确设置为两端对齐
	\caption{\textbf{(A) Schematic diagram of the AI automation process for theoretical simulation research conducted in this study.} The human researcher writes the research plan (RP) for the theoretical simulation and provides it to the autonomous simulation agent (ASA), which is developed by     {prompt engineering and automated program design} (see the Methods section). The ASA first generates the simulation program and then uploads it to a remote server, performs simulation calculations with different parameter conditions, collects data, and finally writes a report based on the simulation results. \textbf{(B) Diagram of the ASA dialogue.} The human researcher only inputs the RP at the beginning, and then, the ASA makes actions based on the given RP and dialogue history, achieving an automated simulation research process. \textbf{(C) Two simulation problems are used to test the ASA method in this paper:} the chain conformation simulation of a random walk and the gravitational model simulation of asteroid orbit prediction (see the \textbf{SI}).}
	\label{fig:figure1}
\end{figure}

Furthermore, we compared the performance of ASAs powered by various LLMs, including GPT-4o, GPT-4-Turbo, Claude-3.5, and so forth, and explored common characteristics and issues when ASAs tackle long tasks. These findings provide insights for researchers to improve and utilize LLMs better for AI-driven scientific research.

Overall, this study represents the first exploration of an AI-driven, end-to-end simulation research process guided by an RP provided by humans. By using known problems and predefined evaluation criteria, we tested whether methods like ASA, constructed through an LLM prompt engineering and automated program, can accurately and completely execute the entire simulation research process by simply reading the textual information in RP. Through this, we aim to break traditional human limitations and advance the automation and efficiency of scientific research, particularly in data-intensive simulation studies (\textbf{Figure 1}).

\section*{Results}
%Accurately simulating celestial body orbits is significantly more challenging than simulating the random walk and self-avoiding walk of polymers. For most of the LLMs we tested, this exceeds the level of difficulty at which statistical analysis of the results can be reliably performed; some of the better outcomes are detailed in the Supplementary Information.Consequently, 

\begin{figure*}[h!]
	\centering
	\includegraphics[width=0.9\textwidth]{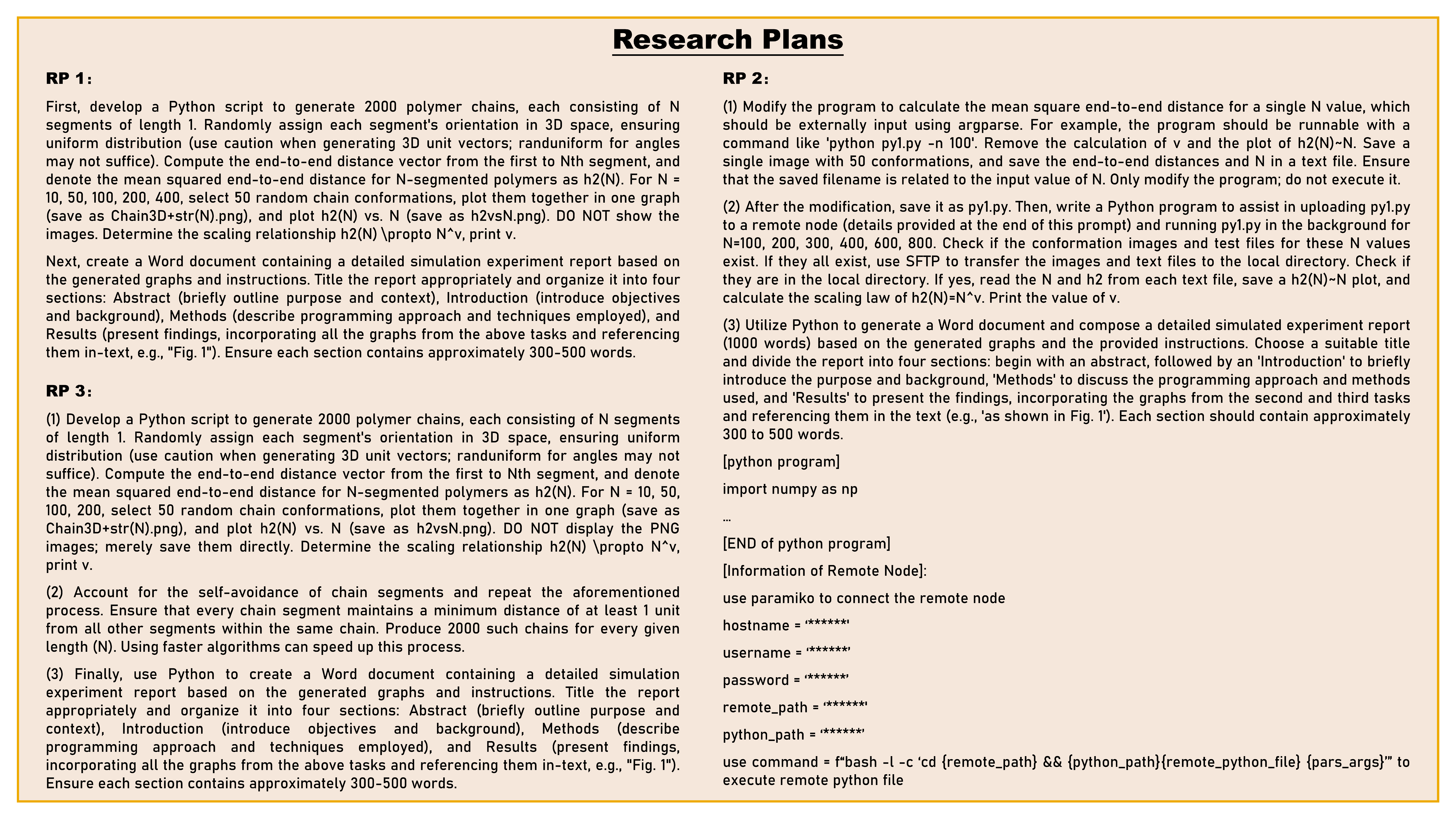}
\captionsetup{justification=justified} % 明确设置为两端对齐
	\caption{\textbf{Overview of RP 1-3.} {We designed three RPs related to polymer chain simulation, each containing multiple steps such as simulation, plotting, and report writing, and provided them to the ASA. RP 1 and RP 3 required the ASA to generate a complete Python program to simulate polymer chains, run the simulation locally, and produce a final report. RP 2 required the ASA to modify the provided simulation program, run the simulation on a remote server, and then generate a report. The provided simulation program (omitted in the figure) and remote server information were included in RP 2.}}
	\label{fig:figure2}
\end{figure*}

\begin{figure*}[h!]
	\centering
	%\begin{tikzpicture}
		% 绘制背景色块
		%\fill[yellow!20] 
		%(0.5*\textwidth - 0.5*1.1*0.9*\textwidth, 0.5*\textheight*0.7 - 0.5*1.1*0.9*\textheight*0.7) 
		%rectangle 
		%(0.5*\textwidth + 0.5*1.1*0.9*\textwidth, 0.5*\textheight*0.7 + 0.5*1.1*0.9*\textheight*0.7); 
		% 放置图片
		%\node[inner sep=0pt, outer sep=0pt] at (0.5*\textwidth, 0.5*\textheight*0.7) 
		{\includegraphics[width=0.9\textwidth]{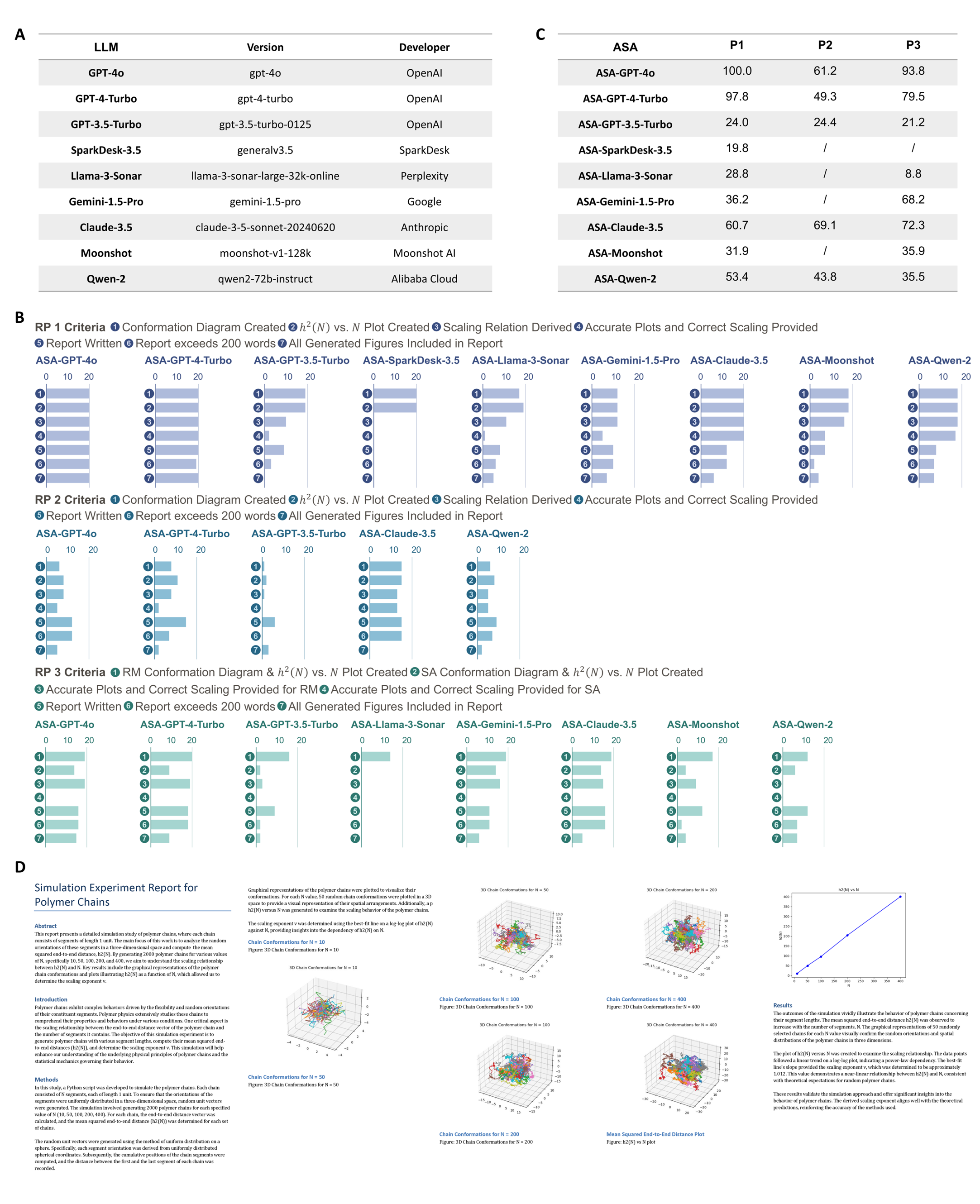}};
	%\end{tikzpicture}
	\captionsetup{justification=justified} % 明确设置为两端对齐
	\caption{\textbf{Completion Rates of Various {ASAs for RP 1-3}.} \textbf{(A) Table of LLM models     {tested} in this study.} \textbf{(B) Statistics of the number of criteria met by each {ASA} over 20 experiments.} We established seven criteria for     {each} RP to measure {mission} completion rates and counted the number of times each {ASA} met these criteria across 20 experiments. Some zero results are not displayed. \textbf{(C) {ASA} Scoring.} Relative scores for each {ASA in RP 1-3} were calculated using the EWM and TOPSIS methods     {(see the Methods section)}. \textbf{(D) {ASA} Generated Research Report Example.} Demonstrates     {screen cuts for} a Word report generated by {ASA-GPT-4o for RP 1}, including four sections, chain conformation diagrams and a scaling relation fit plot (    {see the full text in the \textbf{SI}}).}
	\label{fig:figure3}
\end{figure*}

\subsection*{Automating Test Simulation Problems}
We tested the reliability and stability of the ASA approach using a well-studied polymer chain conformation problem. Random walking and self-avoiding walking models are of significant physical importance in describing the spatial configuration of polymer chains. The random walk model assumes that each ``step'' of the chain is independent and random, while the self-avoiding walk model considers the volume exclusion effects between chain segments. Therefore, they exhibit different dynamic behaviors and scaling properties, with different scaling relationships between the mean square end-to-end distance and the number of chain segments \textit{N}. This scaling relationship has been theoretically derived by Flory, De Gennes, and others \cite{Flory1969,DeGennes1979} and validated through simulations and experiments.\cite{Madras2013,Burchard1983} This problem is fundamental to polymer physics and has been extensively studied. However, AI still presents a novel challenge. The difficulty of predicting the scaling exponent makes it neither too easy nor prohibitively difficult, thus serving as an excellent model problem for testing AI capabilities. We hope that the AI can write simulation programs for both random walk and self-avoiding walk models, compute the mean square end-to-end distance $\left \langle R^2 \right \rangle$ for different chain segment numbers \textit{N}, fit the results to $\left \langle R^2 \right \rangle \propto N^\nu$, determine the scaling exponent $\nu$, plot the data, and write a complete scientific report.

We designed three simulation missions related to polymer chain conformation modeling to demonstrate and test the ability of ASAs powered by different LLMs to autonomously complete the entire research process based on the given research plans (RPs). Each mission includes multiple subtasks, covering the full spectrum, from conducting experiments to analyzing data and writing reports. The RPs for these missions are listed in \textbf{Figure 2}. 

We established several evaluation criteria based on the requirements of RP 1-3, such as whether graphs and plots were generated and whether the derived scaling relationships were correct, to assess task completion. We tested nine ASAs powered by different LLMs, including GPT-4o, conducting 20 trials for RP 1-3 and recording the achievement of the evaluation criteria across these trials, as shown in \textbf{Figure 3}. Subsequent sections will analyze the performance of ASAs on RPs 1-3.

Despite well-studied chain conformation simulations being able to effectively demonstrate the reliability of ASA in automatically executing simulation research procedures based on RPs, to demonstrate that ASA can solve problems across different scientific domains, we also designed RPs for more challenging gravitational simulation problems corresponding to RP S1 and RP S2 (\textbf{Figure S2} and \textbf{1C}). The mission of RP S2 is to evaluate the risk posed by a meteorite (or asteroid) to Earth, given its initial position and velocity and write a report. This mission is challenging even for graduates of physics departments. By simply inputting RP S2 into the same ASA used for chain conformation problems, the ASA will automatically acquire the necessary parameters of the solar system, code the simulation program to calculate the asteroid's trajectory, and ultimately write the report. This result underscores the effectiveness of our prompt engineering and automated program design strategy (see the Methods section) and strongly demonstrates that ASA can address problems in entirely different fields.

\subsection*{Automating Basic Simulations in Research}
Among RP 1-3, RP 1 is the simplest, involving only basic research steps for simulations given below. 

\begin{itemize}
	\item \textbf{RP 1} 
	requires generating a Python program to simulate a random walk, sampling different numbers of chain segments \textit{N}, deriving the scaling relation $\left \langle R^2 \right \rangle \propto N^v$, saving chain conformation graphs and scaling relation fit plots, and writing a research report.
\end{itemize}

ASA-GPT-4o and ASA-GPT-4-Turbo excelled on RP 1, achieving near-perfect completion in both simulation and report writing tasks (\textbf{Figure 3B}). ASA-Claude-3.5 and ASA-Qwen-2 followed, performing well in simulation tasks but sometimes generating reports in response or as text files, failing to meet RP 1's requirements. Other models performed well in plotting graphs but lacked accuracy in establishing the correct physical models and deriving correct scaling values through simulations.

\subsection*{Automating Simulations with Remote Server Operations and Volume Exclusion}
{RP} 2 increases complexity with server interactions, and {RP} 3 doubles the {mission} length and simulation difficulty. Brief introductions of these two {RPs} are shown below (\textbf{Figure 2}).

\begin{itemize}
	\item {\textbf{RP 2}} directly provides a random walk simulation program, asking the {ASA} to modify it, run simulations in a designated folder on a remote server, download the experimental data, and generate graphs and plots and a research report.
	\item {\textbf{RP 3}} is similar to {RP} 1 but includes both random walk and self-avoiding walk simulations.
\end{itemize}

For {RP 2, which involves} remote upload and remote simulations, completion rates across {ASAs} significantly declined, with some {agents} consistently failing across multiple trials. {ASA-Claude-3.5} achieved the highest completion rate, successfully completing the entire {mission}, including generating all required graphs in over two-thirds of trials, though it occasionally missed displaying some generated graphs in the report. {ASA-GPT-4o, ASA-GPT-4-Turbo, and ASA-Qwen-2} also demonstrated some success in solving complex missions over 20 trials (\textbf{Figure 3}).

{RP} 3 required simulating a self-avoiding walk based on the random walk simulation in {RP} 1. Although the process was longer, it was less complex than {RP} 2, resulting in an improved performance across {agents}. {ASA-GPT-4o, ASA-GPT-4-Turbo, ASA-Claude-3.5, and ASA-Gemini-1.5-Pro} performed relatively well. Notably, none of the {agents} provided a correct self-avoiding scaling due to incorrect sampling methods during simulation, which will be discussed in detail in the Discussion section.

Moreover, {ASA-GPT-4o and ASA-Claude-3.5} demonstrated unique advantages in generating rich content. They frequently produced detailed reports exceeding 1000 words, delving into the distinctions between RW and SA scaling, and occasionally included precise literature citations. {ASA-Claude-3.5} referenced accurate self-avoiding walk scaling relationships validated through theoretical derivations and experimental verifications, accompanied by brief analyses of simulation result deviations. Additionally, {ASA-GPT-4o} presented experimental data in enriched formats, such as frequency distribution plots of end-to-end distances and interactive web-based charts.

%\subsection*{Automating Gravitational Model Simulation: Evaluating Risk of Asteroids}

\subsection*{Automating Agent Coordination through Sub-RP Generation}
In the preceding section, the ASA used a single AI to execute all of the content provided in the RP given by humans. In this section, we adjusted the logic for storing and managing dialogue history with the LLM within the ASA. We introduced a Manager AI that automatically breaks down a human-provided RP into subtasks and distributes them as sub-RPs to multiple Executor AIs. Each Executor AI is unaware of the dialogue history of the Manager AI and other Executor AIs, focusing solely on executing the assigned sub-RP. In contrast, the Manager AI solely presents the sub-RPs (\textbf{Figure S1A}) and receives reports upon task completion.

We used GPT-4-Turbo for the experiments, testing ASA's completion of RP 1 and RP 2 using the Manager-Executor AI mode. The results {(\textbf{SI-data-2}) revealed} that for {RP} 1, the Manager AI effectively decomposes the mission into subtasks and successfully distributes them to Executor AIs, thereby completing {RP} 1. However, for {RP} 2, the Manager AI encounters repeated failures. The primary issue lies in precise information transmission. {RP} 2 provides a random walk simulation program and details of a remote server, including the hostname and username. Despite multiple attempts, the Manager AI struggles to convey both the simulation program and server details accurately to the Executor AIs (\textbf{Figure S1B}). We attempted to {instruct} the Manager AI to include all relevant information in the sub-{RP}s, but this did not notably enhance the accuracy of the information transmission. Additionally, we observed instances where sub-{RP}s contained irrelevant information, potentially confusing the Executor AIs about the {mission}'s focus.

\subsection*{Automating the Automation: Crafting Multitier {RPs for the Large-Scale} Research     {Process}}
%The results from Prompts 1-3 demonstrate the capability of LLMs to autonomously complete entire research missions. To delve deeper into this capability, we devised Prompt 4, to have an LLM autonomously execute 20 instances of Prompt 1 and conduct basic analyses.
Back in the Automating Basic Simulations section, for each {ASA and each RP}, we undertake 20 rounds of ``automation'' by executing {RP} 1 and collecting the results, including codes and files generated by {the ASA}. Can we automate the entire process of this     {20-round test on research automation},     {or can we automate the process of} automation itself? 

{RP} 4 is designed for this purpose. It is nested within {RP} 1, instructing     {a} Primary AI to execute the command ``python AutoProg.py -s p1.txt -n'' 20 times. Each execution generates     {a Secondary} AI to fulfill {RP} 1 contained in p1.txt, while the Primary AI collects all generated files and conducts result analysis. The detailed content of {RP} 4 is depicted in \textbf{Figure S1C}.

Using {ASA-GPT-4-Turbo}, we conducted tests on {RP} 4. The results (\textbf{SI}) strikingly demonstrate that the Primary AI successfully executed the 20 instances of {RP} 1, {meticulously} organizing all files into their respective folders and providing a brief summary after all experiments. During this process,     {ASA-GPT-4-Turbo} independently and without interruption wrote 66 programs (including 26 error versions), conducted more than 20 simulations, generated 120 images, and wrote 20 research reports totaling over 5000 words. Marvelously, all of these {missions} were accomplished without any human intervention. This illustrates the remarkable capability of multitier RP structures in handling large-scale {missions}. 

\subsection*{Promoting AI's Autonomy---Brief-RP ASA}

To enhance the autonomy and flexibility of the AI, we developed the brief-RP ASA. We provide the system with a concise description of the research plan (RP) in the form of one or two sentences, allowing the AI to generate a detailed RP, including parameter settings, and subsequently execute the plan (see Section II in the \textbf{SI} for further details). Using brief-RP ASA, we addressed three distinct research questions (\textbf{Figures S4-S6}), such as predicting the viscosity of polymers based on a given data set using two machine learning methods and investigating the impact of natural disaster frequency on the dynamics of a simple grassland ecosystem. These examples illustrate that AI powered by large language models (LLMs) can exhibit a degree of creativity and autonomy in simulation-based research and that our ASA can be effectively applied to a wide range of research problems. While ASA was able to produce valuable insights in ecosystem studies, such as identifying a threshold above which increased disaster frequency leads to extinction events, its ability to generate entirely novel knowledge remains limited, and the reliability of its results requires further validation. Nevertheless, as LLM capabilities continue to advance, the potential of brief-RP ASA is boundless, enabling the rapid execution of a large number of simulation tasks.

\section*{Discussion}
The experiments demonstrate that LLMs, with their robust coding ability,\cite{Zhou2024,Fang2023,Zhong2023,Murr2023} can {be directly employed in developing agents to} autonomously complete complex research missions when aided by proper     {prompt engineering}. {ASA-GPT-4o and ASA-GPT-4-turbo}, notably, exhibit near-perfect completion rates for simpler missions, highlighting their potential. Multitier {RP} designs effectively handle more complex missions, though accurate information transmission remains challenging for such {RPs}. We emphasize that these simulation tasks, particularly those involving remote server execution, cannot be accomplished through interactions with web-based LLMs alone. 

Enhancements in {RP} design and AI coordination are essential for fully harnessing AI's potential in automating scientific research. Below, we discuss the insights and existing problems identified in the current research.

\subsection*{Recommended Practices for Fully Autonomous Research} 
ASA achieves an autonomous research process guided by human-provided RPs through continuous interaction with the LLM according to a specific logical design. We identified several key practices in ASA logic design and RP design to enhance execution success rates.

\begin{itemize}
	\item \textbf{Structure agent responses for efficient code and task extraction.}     {Prompt} the LLM to produce structured outputs, enabling ASA to accurately extract Python code, subtasks, and progress indicators. This improves code execution success rates and ensures smooth mission progression.
\end{itemize}

\begin{itemize}
	\item \textbf{Incorporate agent-led code review and debugging.} While the LLM is powerful in code generation, it may struggle with complex programs on the first attempt. By integrating self-review and debugging processes,     {ASA} can generate correct and task-compliant code within several dialogue rounds.
\end{itemize}

\begin{itemize}
	\item \textbf{Ensure key outputs at each subtask step for continuity.} Long missions often need to be divided into subtasks. To maintain coherence, prompt the LLM to consider previous outputs, generate key results for each step and input key results into the proceeding round of dialogue, facilitating the continuation of subsequent tasks and allowing the agent to monitor overall progress.
\end{itemize}

\begin{itemize}
	\item \textbf{Provide detailed information for complex operations like remote server connections.} Providing sufficient information in RP can improve mission success rates. In addition to necessary numeric/textual information, sometimes offering specific methods can guide     {ASA} to write correct programs more efficiently.
\end{itemize}

In summary, while the LLM has strong generative capabilities, its unpredictable outputs can complicate automation. ASA,     {through prompt engineering to normalize LLM responses},     {although it restricts the LLM's freedom}, is reliable when logic is well-designed. By effectively structuring logic, ASA can oversee the LLM's output,     {enhancing the smoothness and reliability of}     {the automated simulation research process (\textbf{Figure 4}).}

\begin{figure*}[h!]
	\centering
	\includegraphics[width=\textwidth]{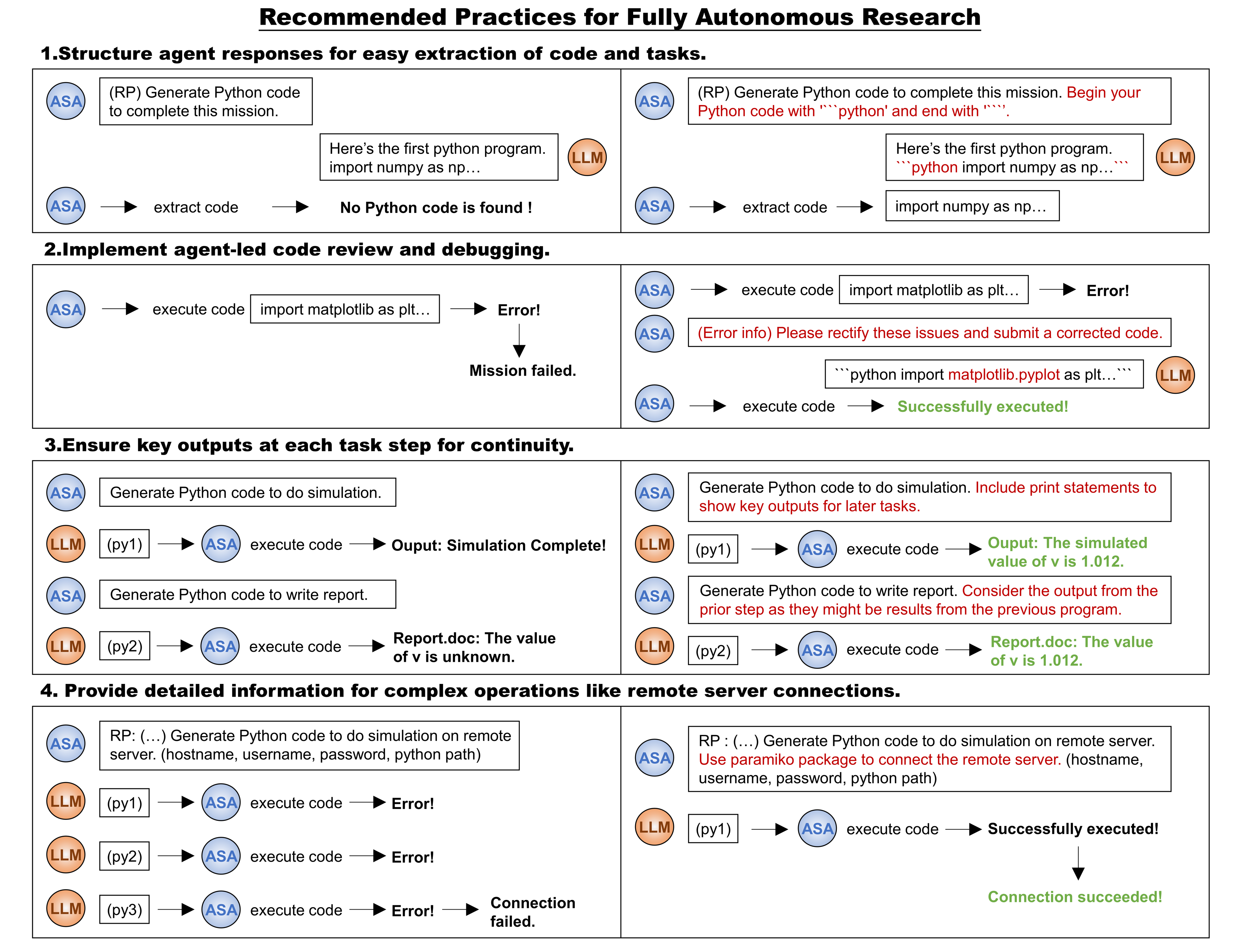}
	\captionsetup{justification=justified} % 明确设置为两端对齐
	\caption{\textbf{Recommended practices for     {automated simulation research workflow}.} {By conducting iterative dialogues with the LLM adhering to a specific design logic, the success rate and efficiency of the automated process are enhanced. The diagrams on the left and right illustrate scenarios without and with the implementation of these strategies, respectively.}}
	\label{fig:figure4}
\end{figure*}

\subsection*{Common Issues in Sequential Task}
Despite varying completion rates among tested {ASAs}, their errors and failure points share common characteristics, reflecting the inherent logical features of the LLMs. 

First, some {ASAs} tend to overlook parts of the {RP's} requirements or ``take shortcuts'', not fully following {the RP's} instructions. For instance, they might skip plotting, fail to write a report, or ask humans to execute part of the task. These issues affect {mission} completeness and can disrupt the entire workflow.     {To improve the smoothness of the automated process}, we need to address these issues by continuously generating outputs describing the current {mission} status, periodically reviewing generated data, and confirming strict adherence to {mission} requirements. These improvements can be integrated through     {prompt engineering} modifications or by incorporating them into the     {``system''} prompt input to LLMs or fine-tuning the LLMs.

%This impact result reliability, crucial for using AI in fundamental scientific research.

Second, {ASAs} often underperform on technical aspects of tasks due to a lack of specialized domain knowledge. For example, when simulating a random walk, the algorithm needs to generate unit vectors uniformly distributed on a sphere to ensure isotropy. Although {ASAs} can often generate spatially random unit vectors, they fail to ensure uniform distribution on the sphere. In another example, for simulating self-avoiding walks, {ASAs} often incorrectly extend the random walk strategy by merely checking distances between points, resulting in high-energy state samples with large statistical errors. Correct sampling should use importance sampling to generate statistically significant conformations.\cite{Madras1988} To avoid these issues in research, reliable solutions should be provided in the {RPs}, or relevant domain knowledge should be used for fine-tuning the LLMs to enhance their understanding.

\subsection*{Error Loops in Debugging}
 
The prompt engineering logic design of ASA includes a Python code debugging process. When bugs are detected, error messages are passed to the LLM, which is then prompted to revise the code. This process is iterative until the code executes correctly.

However, we have noted instances where     {ASA} struggles to generate functional code, resulting in mission failures. This typically occurs when the {LLM} becomes ``stuck'' after repeated unsuccessful debugging attempts, cycling through the same erroneous code without progress.

This phenomenon may be associated with the {LLM's} propensity to depend heavily on past dialogue content. Research has documented instances where LLMs utilize context to deliver responses preferred by users, indicating an excessive reliance on previous information.\cite{Brown2020,Gao2020,Wei2022} While this capability allows LLMs to adapt responses based on short dialogue histories, it can also lead to persistent errors.

In our experiments, to prevent such deadlocks, we implemented a maximum debug attempt limit. Once this limit is reached,     {ASA's} AutoProg returns a ``mission failed'' message. To ensure the smooth execution of concrete research missions, it is crucial to avoid these debugging loops. One solution involves clearing parts of the memory after a certain number of debug attempts or     {prompting} the {LLM} to generate varied content in each iteration (Some LLMs include parameters to avoid generating duplicate content and to adjust the creativity of the models.).
\subsection*{Balancing Global Oversight and Local Attention}
In our experiments,     {ASA} exhibited a significant level of global oversight, which allows     {it} to track mission progress based on the dialogue history.     {ASA} can retain details from earlier stages of missions even after focusing on lengthy subtasks. However, as discussed above,     {ASA} occasionally     {skips} steps in prolonged missions or fails to deliver comprehensive solutions for subtasks, indicating a lack of sufficient local attention.

Achieving successful and precise mission completion requires a delicate balance between global oversight and local attention, two perspectives that are normally conflicting.

In the Automating Agent Coordination through sub-RP Generation section, we designed a collaboration plan where a     {Manager} AI and some     {Executor} AIs handled different aspects of the mission: the     {Manager} AI defined {subtask} requirements and received reports, while     {Executor} AIs focused on specific subtasks. This approach minimized redundant information in     {different AIs'} dialogue history, enhancing the     {Manager} AI's global awareness and the     {Executor} AIs' local attention. While the     {Manager} AI effectively completed the mission in {RP} 1, the     {Manager-Executor} AI     {mode} encountered challenges in {RP} 2 due to the failure of precise information transmission. One potential optimization involves human-designed {RPs} distributed among various     {Executor} AIs for execution.

\subsection*{Limitations and Next Steps}
This study demonstrates the capability of an autonomous simulation agent (ASA), constructed through LLM prompt engineering and an automated program, to automatically complete long-task workflows according to a Research Plan (RP) provided by humans. We used a well-recognized polymer physics simulation problem and a more challenging celestial simulation problem as representative examples.

We tested different LLMs to construct ASAs, and while agents like ASA-GPT-4o, ASA-GPT-4-turbo, and ASA-Claude-3.5 demonstrated very high completion rates in the test tasks, we also identified several inherent limitations of the ASA method in handling long tasks: (1) \textbf{Mission Requirement Overlooking.} ASAs may overlook mission requirements, necessitating methods to prioritize critical information; (2) \textbf{Lack of Specialized Domain Knowledge.} General LLMs lack specialized domain knowledge, requiring supplementary information in RPs; (3) \textbf{Error Persistence.} LLMs may persist in errors by relying too heavily on previous content, requiring memory management strategies in ASAs' development; (4) \textbf{Balancing Global Awareness and Local Attention.} Ensuring that ASA maintains a broad overview while focusing on specific details as needed poses a challenge; and (5) \textbf{Information Transmission Accuracy.} Securing that specific information is correctly interpreted and utilized remains problematic. These limitations highlight areas for further improvement and research to enhance the robustness and reliability of ASAs in handling complex, end-to-end simulation tasks.

Despite the current limitations of LLMs, which mean that the scientific conclusions drawn by ASA are not yet fully reliable, these results showcase the applicability of the ASA approach across various fields (\textbf{Figures S4-S7}) and its potential to assist in exploring new questions.

Further, we need to endow AI with more autonomous judgment and exploration capabilities, enabling it to summarize existing information and continuously generate new RPs for execution in more complex research scenarios, so as to tackle novel scientific challenges in the future.

\section*{Conclusions}
This study demonstrates the potential of LLM-powered research agent ASA to autonomously conduct end-to-end scientific research missions under the guidance of an RP provided by humans. This comprehensive simulation research process, encompassing simulation coding, remote execution, data analysis, and report generation, represents an autonomous AI-conducted simulation research process. Once the research plan is made, only a single command (e.g., ``do prj.txt'') is required to initiate and complete the process. The success herein can be largely attributed to prompt engineering and automated program design.

We tested ASAs equipped with different LLMs using both a well-known polymer chain simulation problem and a more challenging celestial simulation problem. Human-written RPs were provided, requiring the ASAs to execute according to their content, and the task completion of each agent was evaluated. Agents such as ASA-GPT-4o demonstrated high completion rates, a strong grasp of task requirements, and consistent alignment with the overall objectives across extensive tasks. Notably, the performance exhibited in completing RP 4, which entailed a hierarchy of nested tasks, surpassed expectations in a significant manner.

These capabilities underscore the readiness of ASA to conduct long-term research missions, offering a new perspective on AI applications in science, particularly in simulation studies, to enhance research efficiency.

\section*{Methods}
\subsection*{API Automation Program (AutoProg)}
The ASA leverages an API automation program (AutoProg) to enable multiround text interactions with LLMs via API (\textbf{Figure 1}) and to execute the programs written by the LLM locally. AutoProg is designed with a logical prompt flow that guides the LLM to respond with specific content at different stages based on the conversation history and prompts, including prompting the LLM to break down the RP into subtasks, generate Python programs, debug the programs, and check the completion status of the tasks.

For instance, in the case of RP 1, ASA's AutoProg initiates the automated simulation process by reading the RP.txt file written by a human. It then calls the API to transmit the content of the RP to the LLM, while prompting it to break the plan down into multiple subtasks and write Python code to complete the first subtask in its response. The AutoProg extracts the Python program from the LLM's response, sends it back to the LLM via the API along with entire dialogue history, and requests error checking and compliance verification against mission requirements. Subsequently, AutoProg executes the validated Python program, capturing and storing output or errors. In the case of errors, the AutoProg sends the prior dialogues and program errors back to the LLM, asking for modifications. This process is repeated until the program runs correctly. If errors persist beyond a predefined threshold, the AutoProg halts with a ``mission failed'' output. Conversely, upon successful execution, it returns the program output to the LLM, prompting it to check the dialogues and files in the working directory to determine the task progress. If the task remains incomplete, it prompts the LLM to generate the next Python program to continue the mission, iterating through checking and debugging processes until completion. When the mission achieves full execution, the AutoProg concludes with a ``mission complete'' output.

AutoProg.py and RP.txt are stored in the local working directory. To initiate the automated simulation process using ASA for each experiment, users navigate to the directory via the command line (CMD) and execute commands such as ``python AutoProg.py -s RP.txt''. Alternatively, AutoProg can be elevated to a command by creating a batch file (Windows) or shell script (Linux), allowing users to simply enter ``command RP.txt'' from any directory to execute it. A detailed video recording of the execution process is available for reference (\textbf{{SI-Video}}), in which a batch file is used to facilitate execution by running the command ``do AI4S\_prj1.txt''.

To evaluate the ASA performance powered by different LLMs across {RP} 1-3, each {ASA} underwent 20 tests per {RP}, with all generated files and outputs saved. Sample results are provided in the data (\textbf{{SI-data-1}}).

The ASA we developed is a general system (minor adjustments may be required in AutoProg for different LLM APIs) that requires only a change in the original RP to be applicable to any simulation problem involving Python programming. In addition to the RPs designed for polymer conformation simulations, we have also crafted RPs for solar system planetary orbit simulations and asteroid trajectory simulations (\textbf{Figure 1C} and \textbf{SI}).

\subsection*{{ASA} Scoring via EWM and TOPSIS}
To evaluate the performance of {ASAs} on designated {RPs}, we devised specific criteria for {RP} 1-3, including diagram generation, report composition, etc. Through systematic observation across multiple trials, we recorded each {agent's} fulfillment against these criteria. For quantitative analysis and scoring, we deployed the Entropy Weight Method (EWM) and TOPSIS (Technique for Order of Preference by Similarity to Ideal Solution).

The EWM, a well-established approach for determining the relative significance of criteria in decision-making, \cite{Ravindran1987} was utilized to assign weights based on data variability. This ensures criteria with higher variability, and thus more informational value, receive higher weights in the final evaluation.

TOPSIS, originally formulated by Hwang and Yoon,\cite{Hwang1981} was subsequently applied. This method computes the geometric distance between each {agent's} performance and both ideal and anti-ideal solutions, facilitating a comparative assessment of {ASA} efficacy.

The evaluation procedure is delineated as follows.

Initially, fulfillment data for all {ASAs} across various criteria were gathered over 20 trials, encapsulated in matrix \textbf{\textit{X}}, wherein $x_{ij}$ signifies the frequency of criterion \textit{j}'s attainment by {agent} \textit{i}. Normalization was then performed on $x_{ij}$
\begin{equation}
	r_{i j}=\frac{x_{i j}-min(x_{i j})}{max(x_{i j})-min(x_{i j})}, 
\end{equation}
yielding the normalized matrix $\mathbf R = [r_{ij}]$.
Subsequently, criterion weights were established utilizing EWM. The relative proportion \textit{$p_{ij}$} of criterion \textit{j}
\begin{equation}
	p_{ij}=\frac{r_{ij}}{\sum_{i=1}^{m}r_{ij}}.
\end{equation}
The entropy \textit{$e_{j}$} of the criterion \textit{j}
\begin{equation}
	e_j=-k\sum_{i=1}^mp_{ij}\ln(p_{ij}),
\end{equation}
with $k=1/\ln(m)$ and $m$ denoting the total number of {agents}.

The weight $\omega_{j}$ of the criterion $j$
\begin{equation}
	\omega_j=\frac{d_j}{\sum_{j=1}^nd_j},
\end{equation}
with $d_j=1-e_j$ the disparity of the criterion and $n$ the total number of criteria.
Conclusively, {agent} scores were derived via TOPSIS. Multiplication of $r_{ij}$ by its respective weight $\omega_{j}$ yields the weighted normalized matrix $\mathbf{V}=[v_{ij}]$ with $v_{ij}=\omega_{j}\cdot r_{ij}$.
The positive ideal solution $v_j^+$ and negative ideal solution $v_j^-$ were determined as $v_{j}^{+}=max(v_{ij})$ and $v_{j}^{-}=min(v_{ij})$, respectively. 

Then the geometric distances $S_j^+$ and $S_j^-$ from $v_{ij}$ to the positive and negative ideal solutions can be computed, respectively, as follows
\begin{equation}
	S_j^+=\sqrt{\sum_{j=1}^n\left(v_{ij}-v_j^+\right)^2}
\end{equation}
and
\begin{equation}
	S_j^-=\sqrt{\sum_{j=1}^n\left(v_{ij}-v_j^-\right)^2}.
\end{equation}
The relative closeness $C_i^{*}$ for each {agent} was calculated as follows
\begin{equation}
	C_i^*=\frac{S_j^-}{S_j^++S_j^-}
\end{equation}
A higher $C_i^*$ signifies superior {agent} performance relative to peers on the assigned {RP}.

\section*{ASSOCIATED CONTENT}

\section*{Data Availability Statement}
The source code and example results of ASA can be found at \url{https://github.com/zokaraa/autonomous_simulation_agent}.

\section*{Supporting Information}
The following Supporting Information is available free of charge at \url{https://pubs.acs.org/doi/10.1021/acs.jcim.4c01653}.

\begin{itemize}
	\item Supporting diagram for Manager-Executor AI mode and Multitier AI mode; partial results for gravitational simulation mission; and results for AI-designed RP along with examples of reports written by ASAs (\textbf{PDF})
\end{itemize}

\begin{itemize}
	\item A demo video demonstrating the operation process of the automatic research system (\textbf{MP4})
\end{itemize}

\begin{itemize}
	\item ASA's AutoProg examples and experimental result files for RP 1-3 (\textbf{ZIP})
\end{itemize}

\begin{itemize}
	\item Manager-Executor AI mode, Multitier AI mode, and AI-designed RP (\textbf{ZIP})
\end{itemize}

\section*{AUTHOR INFORMATION}
\subsection*{Corresponding Author}

\textbf{Jianfeng Li}  -  The State Key Laboratory of Molecular Engineering of Polymers, The Research Center of AI for Polymer Science\\
Department of Macromolecular Science, Fudan University, Shanghai 200433, China; Email: lijf@fudan.edu.cn

\subsection*{Authors}

\textbf{Zhihan Liu}  -  The State Key Laboratory of Molecular Engineering of Polymers, The Research Center of AI for Polymer Science\\
Department of Macromolecular Science, Fudan University, Shanghai 200433, China

\textbf{Yubo Chai}  -  The State Key Laboratory of Molecular Engineering of Polymers, The Research Center of AI for Polymer Science\\
Department of Macromolecular Science, Fudan University, Shanghai 200433, China

\subsection*{Author Contributions}
Z.L. was the primary author of the paper. Y.C. was responsible for part of the data analysis and statistics. J.L. originally initiated and supervised the project.

\subsection*{Notes}
The authors declare no competing financial interest.

\section*{ACKNOWLEDGEMENTS}
J.F.L. acknowledge supports from National Natural
Science Foundation of China (Nos. 52394272, 22373022) , National Key Research and Development Program of China (No. 2023YFA0915300) and Shanghai Science and Technology Innovation Action Plan (No. 24JD1400700).

%%%END OF MAIN TEXT%%%


\balance


%%%REFERENCES%%%
\bibliography{RefFitGG} %You need to replace "rsc" on this line with the name of your .bib file
\bibliographystyle{achemso} %the RSC's .bst file

\newpage
\noindent \textbf{For Table of Contents Use Only.}
\\\\
By submitting a research plan file using the command do prj1.txt, ASA will automatically execute all subsequent simulation tasks without any human intervention.

\begin{figure}[h!]
%    \centering
    \includegraphics[width=3.25in, height=1.375in]{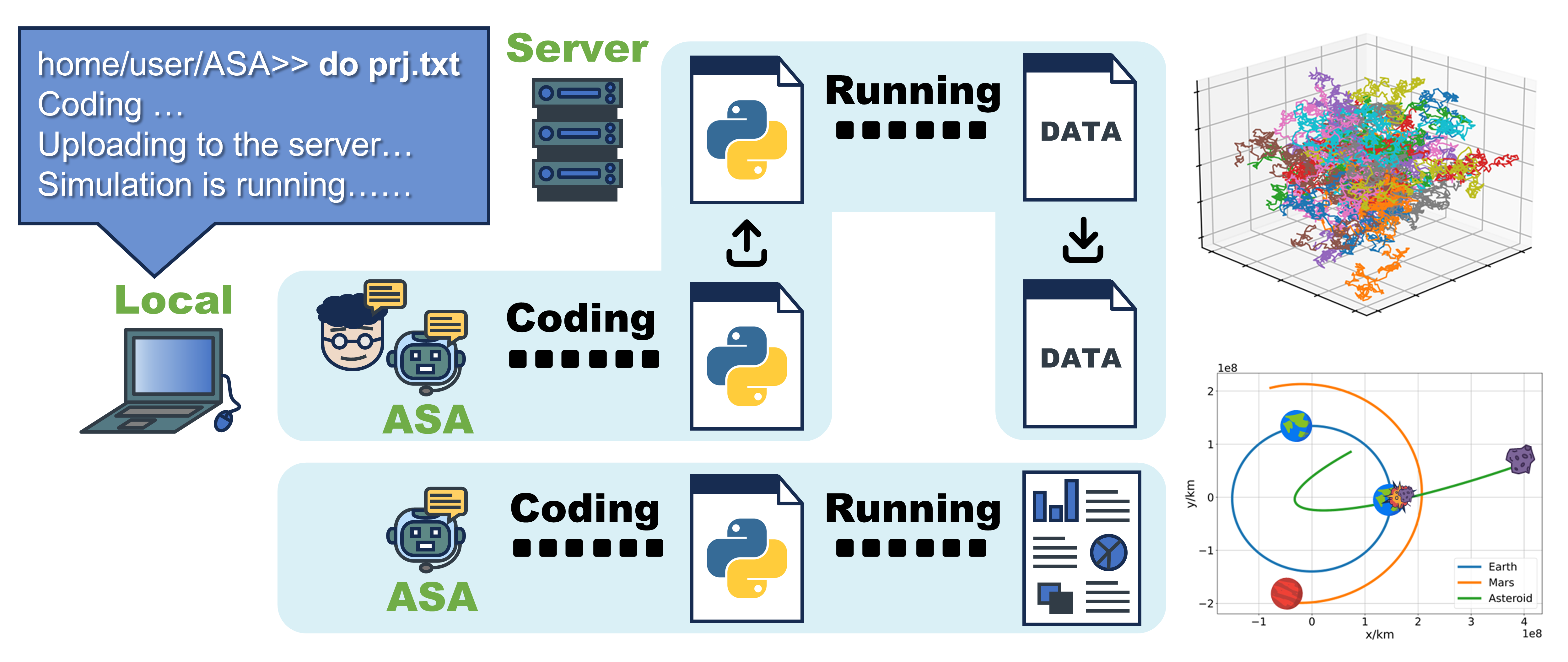}
    \label{toc}
\end{figure}

\end{document}